\documentclass[conference]{IEEEtran}
\IEEEoverridecommandlockouts
\usepackage{cite}
\usepackage{amsmath,amssymb,amsfonts}
\usepackage{algorithmic}
\usepackage{graphicx}
\usepackage{textcomp}
\usepackage{xcolor}
\usepackage[hyphens]{url}
\usepackage{hyperref} 
\def\BibTeX{{\rm B\kern-.05em{\sc i\kern-.025em b}\kern-.08em
    T\kern-.1667em\lower.7ex\hbox{E}\kern-.125emX}}

\usepackage{fancyvrb}
\usepackage{mdframed}
\usepackage{listings}
\usepackage[numbers]{natbib}

\begin{document}

\title{Signal Knowledge Graph}


\author{
\IEEEauthorblockN{Anj Simmons\IEEEauthorrefmark{1}, Rajesh Vasa\IEEEauthorrefmark{1}}
\IEEEauthorblockA{\IEEEauthorrefmark{1}Applied Artificial Intelligence Institute, 
Deakin University, Geelong, Australia\\
Email: \{a.simmons, rajesh.vasa\}@deakin.edu.au}
}

\maketitle

\begin{abstract}
This paper presents an knowledge graph to assist in reasoning over signals for intelligence purposes. We highlight limitations of existing knowledge graphs and reasoning systems for this purpose, using inference of an attack using combined data from microphones, cameras and social media as an example. Rather than acting directly on the received signal, our approach considers attacker behaviour, signal emission, receiver characteristics, and how signals are summarised to support inferring the underlying cause of the signal. 
\end{abstract}


\section{Introduction}

Intelligence agencies are tasked with identifying and quantifying potential threats to provide governments, military and police with the situational awareness needed to defend against them. While intelligence agencies have access to increasing volume, velocity and variety of data at different levels of veracity, making sense of this data to proactively identify threats presents a challenge due to finite human resources available to analyse the data.

To assist humans to analyse data, new techniques are needed to automatically fuse and reason over the data. Ontology based information fusion \cite{Pai2017} is a promising approach for automatically integrating and reasoning about heterogeneous information sources to support situational awareness. This involves constructing a knowledge graph (expressed using concepts from an ontology) describing known information about a scenario, linking it to other knowledge graphs of complementary information, performing inference to derive new facts, and then querying the results for high-level information of interest. However, in this paper we demonstrate that current knowledge graphs lack key information and concepts needed to reason holistically about sensor data.



In \autoref{sec:mot}, we provide an intrusion detection scenario involving microphones, cameras and social media, which is used to motivate the need for human-like reasoning over sensor data. In \autoref{sec:existing-knowledge-graphs}, we identify gaps and limitations of existing knowledge graphs/bases for the purpose of supporting reasoning about the cause of sensor observations. In \autoref{sec:approach}, we propose construction of a signal knowledge graph to support inferring the underlying cause of the signals detected by sensors, which we demonstrate in \autoref{sec:demonstration}. Finally, we outline related work in \autoref{sec:related-work} and conclude in \autoref{sec:conlustion}.

\section{Motivating Scenario}
\label{sec:mot}
As motivation, consider the scenario of an intelligence officer assigned to monitor a building for signs of intruders where an important meeting will take place. To safeguard the building, a network of surveillance devices is set up throughout the building to monitor for suspicious sounds, objects, and behaviours. As there are too many sensors for the officer to monitor manually, and because meeting attendees are concerned about their privacy, each of the devices are configured to first classify the signal on the device (e.g., as the sound of glass shattering, a knife in video camera footage, etc.) then transmit only the classified signal to a centralised surveillance platform.

\begin{figure}[tpb]
    \centering
    \includegraphics[width=\linewidth]{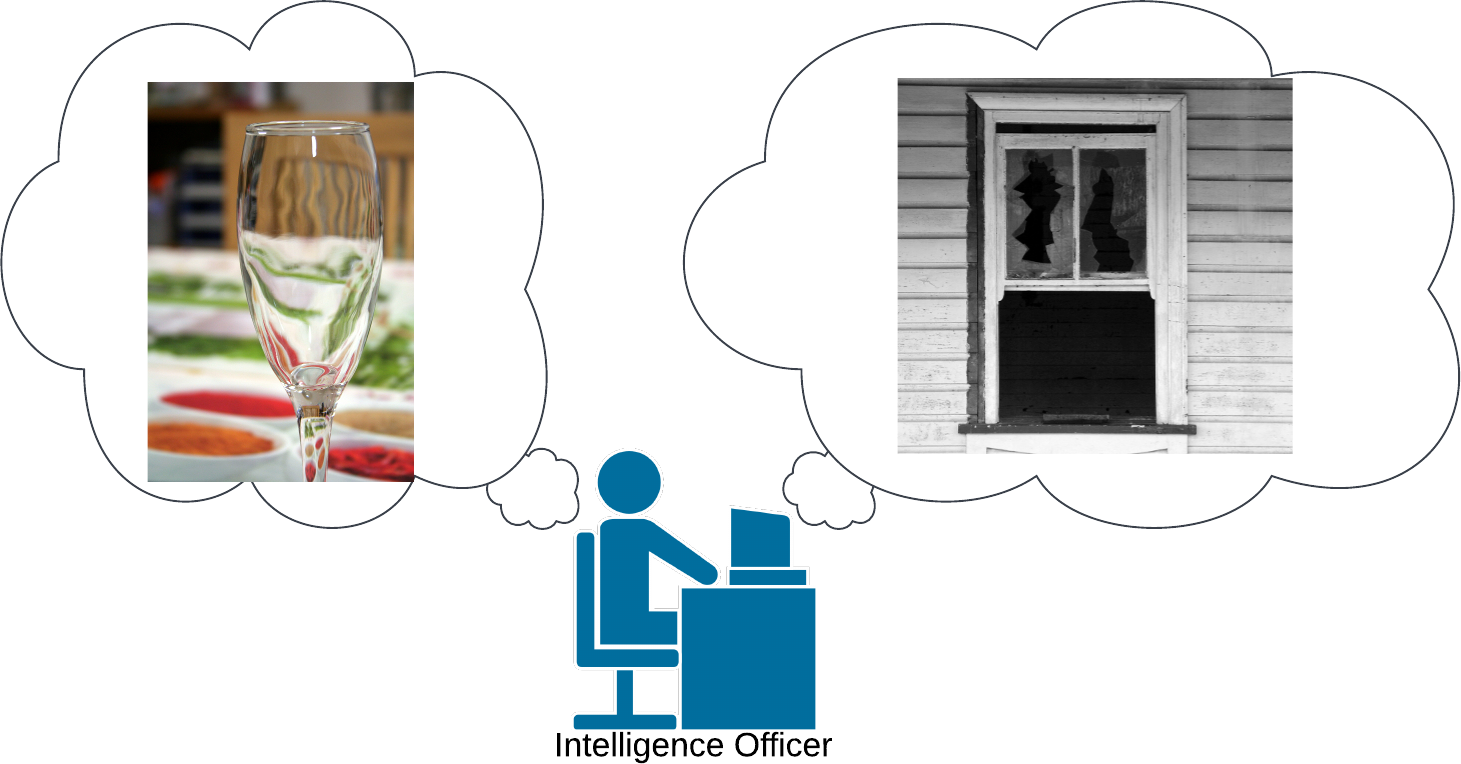}
    \caption{Human intelligence officers consider the possible causes for signals, which allows them to react to subtle indicators of an attack while avoiding false alarms.}
    \label{fig:human}
\end{figure}

\begin{figure}[tpb]
    \centering
    \includegraphics[width=\linewidth]{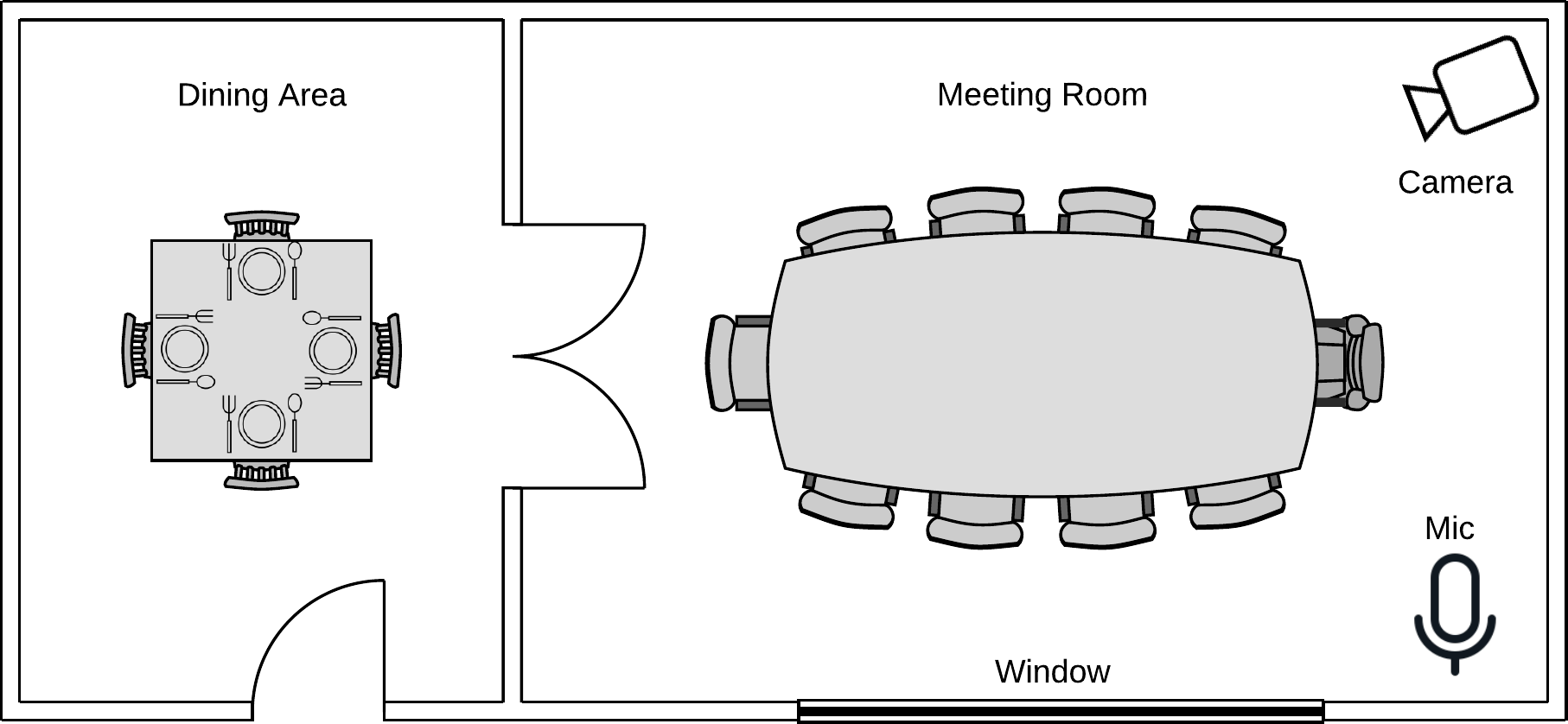}
    \caption{Layout of building for motivating scenario}
    \label{fig:layout}
\end{figure}

The intelligence officer notices that one of the devices has detected a noise classified as `sound of glass', which triggers the thought process shown in \autoref{fig:human}. The intelligence officer looks at a map of the building (\autoref{fig:layout}) where the sensor is located and notices that it is near to a window. The intelligence officer speculates that perhaps an intruder has broken through the window. However, before raising an alarm, the officer considers the possibility that perhaps the noise could have been due to a non-suspicious cause, such as an employee hosting the meeting dropped a tray of glasses in the dining area. As the dining area is far from the microphone and the dining area is unlikely to be busy at the the current time of day, the officer decides that an intrusion is the most likely cause.

This manual approach works for monitoring a particular building, but does not work at scale when an intelligence agency needs to monitor a large number of buildings or spaces. As attacks are rare events, there is unlikely to be enough data for any given building or space to train a machine learning model to recognise attacks. While machine learning can still assist in classifying signals from the sensors as well as detecting anomalies, an anomaly is not necessarily cause for concern. As such, a human intelligence officer still needs to inspect each anomaly to reason about the likelihood that it is an attack rather than a false positive. The intelligence officer can increase thresholds to reduce the number of automated alarms that they need to inspect, but this can result in intrusions going unnoticed. An example is provided in \autoref{lst:hardcode}.

\begin{figure}[htpb]
\begin{lstlisting}[language=Java, caption={Example code for raising an alarm when the probability of a suspicious sound or object exceeds a threshold}, label=lst:hardcode]
for (Sensor mic : sensorData.microphones) {
    if (mic.getProbability("glass") > 0.95) {
       raiseAlarm(mic.location);
    }
}
for (Sensor cam : sensorData.cameras) {
    if (cam.getProbability("knife") > 0.95) {
       raiseAlarm(cam.location);
    }
}
...
\end{lstlisting}
\end{figure}

To address this issue, in this work we propose a knowledge graph to support reasoning over sensor data in a manner closer to that of a human intelligence officer. In contrast to anomaly detection which just detects something is out of the ordinary, our approach explicitly considers the potential causes for signals to reason about the probability that the cause is an attack rather than a rare but non-threatening cause. An example is provided in \autoref{lst:infercode}.

\begin{figure}[htpb]
\begin{lstlisting}[language=Java, caption={Example code for raising an alarm when the inferred probability of an attack exceeds a threshold}, label=lst:infercode, basicstyle=\ttfamily\scriptsize,rulesep=1pt, frame=single, captionpos=b]
Posterior cause = reasoner.inferCause(sensorData, signalKG);
if (cause.getProbability("attack") > 0.5) {
   raiseAlarm(cause.getLocationDistribution("attack"));
}
\end{lstlisting}
\end{figure}




\section{Limitations of Existing Knowledge Graphs}
\label{sec:existing-knowledge-graphs}

There are existing knowledge graphs dedicated to expressing factual knowledge, common sense, and domain specific information. While each of these contain information that may assist in interpreting sensor data, they miss key information needed to support reasoning.

As an example of knowledge graphs focused on factual knowledge, Wikidata is a community maintained knowledge base of over 97 million items. It encodes structured information such as glassware (Q1922981) is made from material (Property:P186) glass (Q11469) and that glass (Q11469) has a density (Property:P2054) between 2,200 and 7,500 gram per cubic centimetre. However, as its focus is on collecting factual information, it lacks common sense information that humans use to reason. For example, it does not include the fact that glassware breaks easily and creates a sound when it does so.

As an example of knowledge graphs focused on common sense information, Concept Net \cite{Speer2017} provides a graph of general knowledge. Concept Net includes some concepts that may assist machines to reason about sensor data. For example, \autoref{lst:conceptnet} shows the relationship in Concept Net between a `break in' and a `chink' sound. However, as reasoning about signals is not the primary purpose of Concept Net, information about the signals produced by an act are limited. Furthermore, as Concept Net focuses on words rather than well-defined concepts, when a single word has multiple meanings it can lead to ambiguities.

\begin{figure}[htpb]
\begin{lstlisting}[caption={Relationship between `break in' and `chink' in Concept Net}, label=lst:conceptnet]
/c/en/break_in /r/MannerOf /c/en/crack
/c/en/chink /r/IsA /c/en/crack
\end{lstlisting}
\end{figure}

Standardised ontologies exist for expressing domain specific information, such as the Semantic Sensor Network (SSN) ontology\footnote{\url{https://www.w3.org/TR/vocab-ssn/}} for describing sensors and their observations, and the Digital Twins Definition Language (DTDL)\footnote{\url{https://github.com/Azure/opendigitaltwins-dtdl/}} describing the physical environments in which sensors are placed. However, reasoning about the underlying cause of sensor observations requires not only knowledge of the sensors and their environment, but also an understanding of the signals they detect and the possible causes of these signals.

To address this gap, in this work we propose construction of a signal knowledge graph encoding the missing information needed to reason about the underlying cause of sensor observations.

\section{Our Approach}
\label{sec:approach}

\begin{figure}[htpb]
    \centering
    \includegraphics[width=0.75\linewidth]{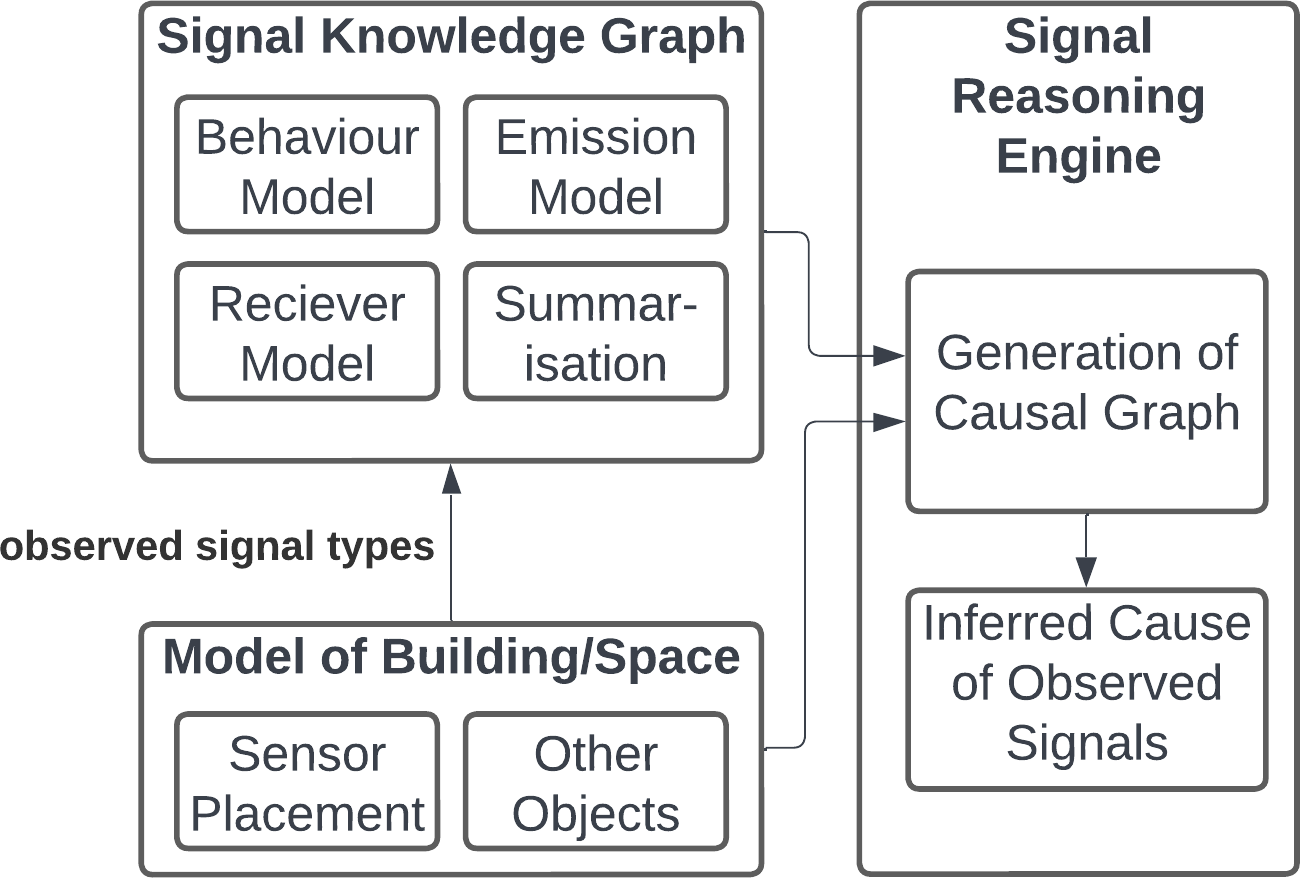}
    \caption{Overview of our approach}
    \label{fig:overview}
\end{figure}

\begin{figure*}[htpb]
    \centering
    \includegraphics[width=0.7\linewidth]{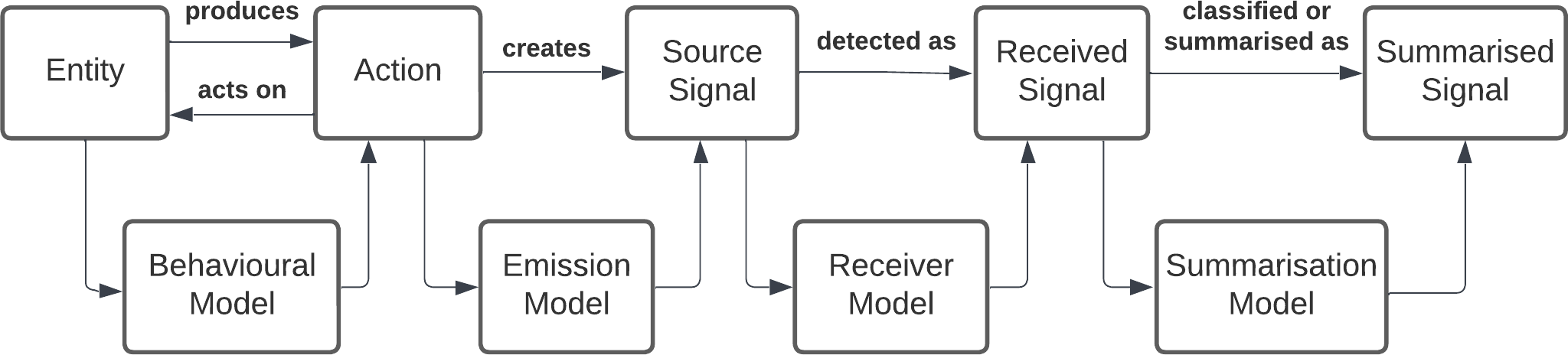}
    \caption{Signal Ontology}
    \label{fig:signal-kg-full}
\end{figure*}

Our approach (\autoref{fig:overview}) involves construction of a signal knowledge graph containing the information needed to reason about different kinds of signals. Together with a model of sensor placement within a building/space (which can be described using existing standards), this can be used to generate a causal graph describing the probability of a observing a set of sensor observations given a set of actions that caused them. Bayesian reasoning is used to infer the cause given sensor observations.


\subsection{Signal Ontology}
\label{sec:signal-ontology}

\autoref{fig:signal-kg-full} presents an ontology describing the high level concepts needed to construct a knowledge graph for reasoning about sensor data. \textit{Entities} (e.g., a person) produce \textit{actions} (e.g., walking), which in turn creates a \textit{source signal} (e.g., the sound of footsteps). The \textit{received signal} detected by the sensor usually reduces in strength with distance and can be distorted by surrounding objects (e.g., a wall between a sound source and the receiver attenuates the sound signal). Finally, the device that receives the signal may perform some pre-processing to \textit{summarise} the signal prior to transmission for efficiency or privacy reasons (e.g., classifying the type of sound, or object recognition in vision systems).

The behaviour, emission, receiver, and summarisation models represent the information needed to reason about this process for each signal type. A formal RDF/OWL specification of our signal ontology and demonstration signal knowledge graph (constructed using concepts in the ontology) are available online\footnote{\url{https://github.com/anjsimmo/signal-knowledge-graph/}}.

\subsection{Signal Knowledge Graph}

Human intelligence officers have an understanding of how signals propagate from the entity producing the action to the receiver and are summarised, at least intuitively even if not familiar with the physical equations governing the system. However machines lack such understanding. Even though computer simulations exist to model particular types of signals at different degrees of fidelity (e.g., ray tracing and photon mapping software for light), the knowledge encoded into the source code of these simulations is not in a form that machines can use to reason. To support machines to reason about signals, we propose constructing an explicit knowledge graph modelling behaviour, emission, receiver, and summarisation. As each type of signal has unique characteristics, for example, a sound-proof window blocks sound but not light, this requires extending the knowledge graph with new information for each type of signal. However, all signals are described using concepts from a common signal ontology (\autoref{sec:signal-ontology}) so that software developed using concepts in the signal ontology can reason about any signal present in the signal knowledge graph.

Observations and sensor placement within a building/space are described using existing standards, such as SSN and DTDL. These can be linked to our signal ontology, e.g., by specifying which type of signal in the signal knowledge graph an observation relates to.

\subsection{Signal Reasoning Engine}

Actions producing signals are probabilistic in nature. Furthermore, even in theoretically deterministic processes, there may be a level of uncertainty involved in practice due to unknowns. The signal reasoning system takes the signal knowledge graph as input, as well as the model of sensor placement within a building/space, and uses this to generate a causal graph. Bayesian reasoning is used to infer the posterior probability of the cause (e.g., an attacker) given sensor observations.

\section{Concept Demonstration}
\label{sec:demonstration}

To demonstrate our approach, in this section we show how our signal ontology can be applied to describe the knowledge needed to reason about sensor observations in the motivating scenario (\autoref{sec:mot}). We model sound/audio signals, vision-based signals, and signals from social sensors. Finally, we show how these different signal types can be integrated as a single causal graph to support inference.

\subsection{Sound/Audio Signals}

\begin{figure}[tpb]
    \centering
    \includegraphics[width=\linewidth]{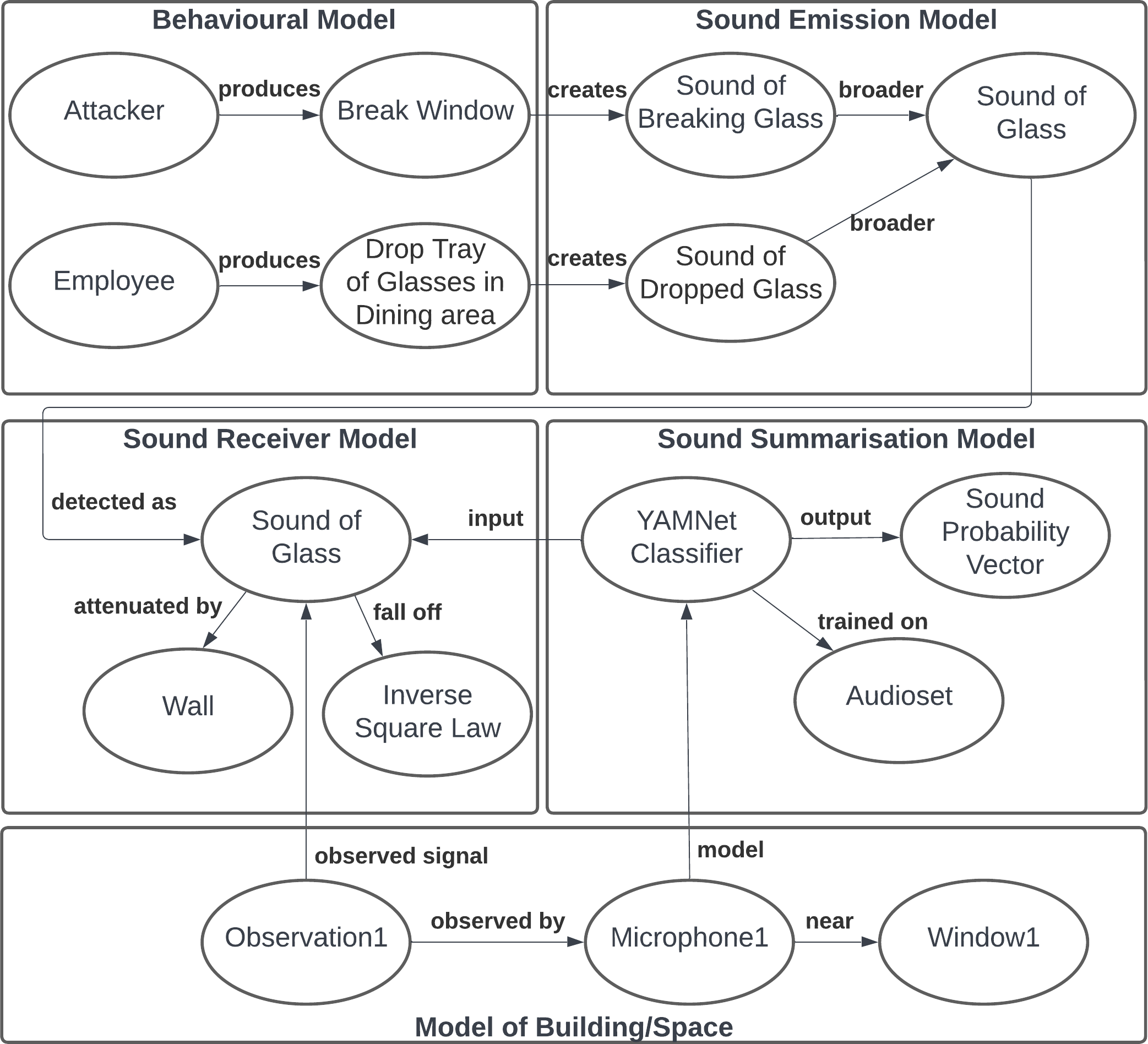}
    \caption{Modelling of sound/audio signals}
    \label{fig:sound-model}
\end{figure}

\autoref{fig:sound-model} shows how concepts in the signal ontology (\autoref{sec:signal-ontology}) can be applied to model sound/audio signals. Attackers and employees are entities that are capable of producing an action, such as breaking a window or dropping a tray of glasses in the dining area. Actions produce signals, in this case, the sound of breaking glass or the sound of dropped glass. To represent the fact that both sounds are similar, we use the Simple Knowledge Organization System (SKOS) \cite{miles2009skos} `broader' property to show that they both belong to the same category. Our receiver model includes the fact sound is attenuated by walls and falls off with distance according to an inverse square law. To model the case in which sounds are classified on the device prior to transmission, we allow for describing different classification models such as YAMNet\footnote{\url{https://tfhub.dev/google/yamnet/1}}, an audio classifier that recognises 521 classes of sounds (including glass) and is trained on the Audio Set \cite{Gemmeke2017} corpus.

\begin{figure}[tpb]
    \centering
    \includegraphics[width=\linewidth]{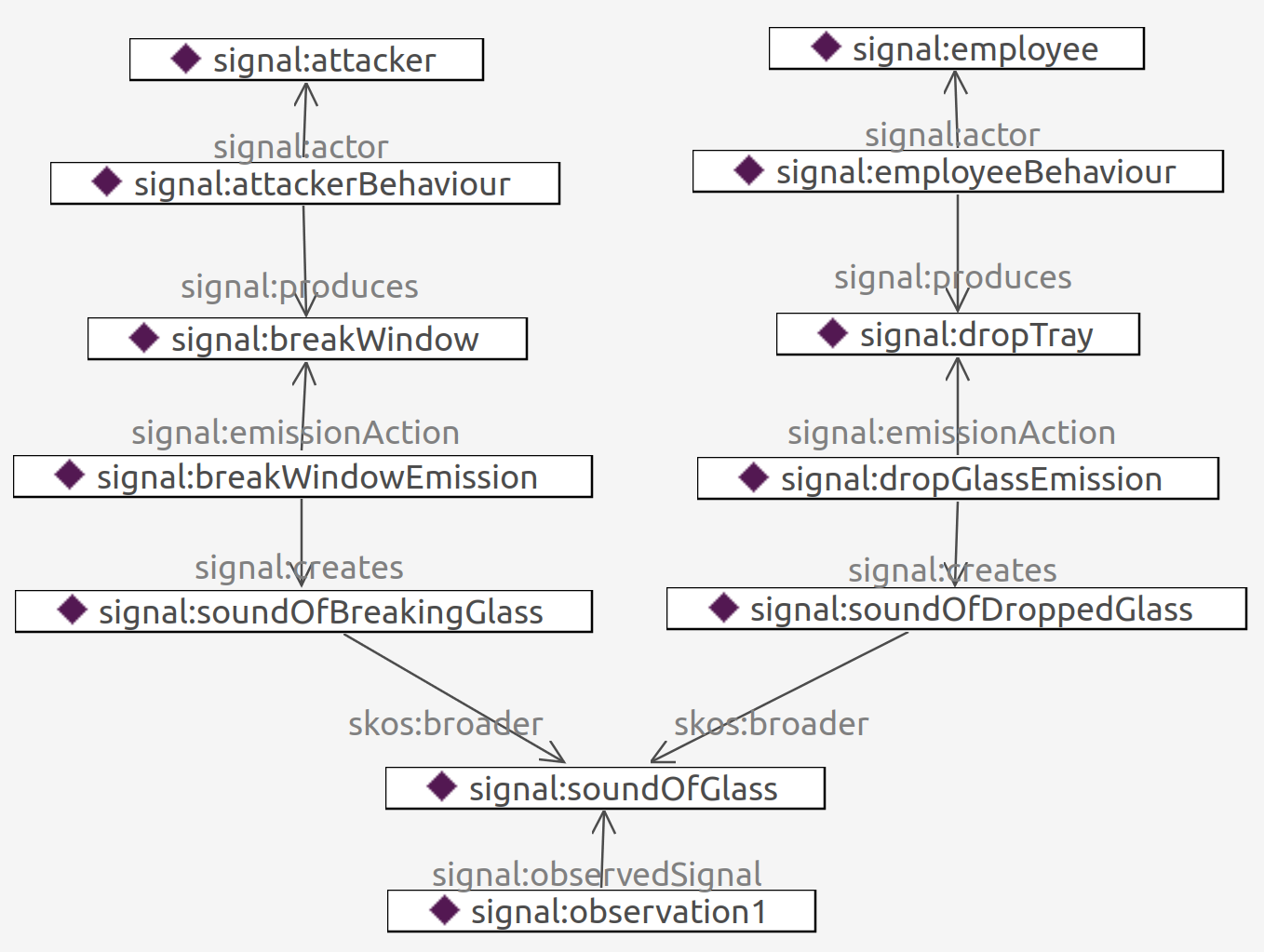}
    \caption{Partial view of knowledge graph created for sound/audio}
    \label{fig:signal-kg-instances}
\end{figure}

We created a formal RDF specification of the knowledge graph, where entities, actions, signal sources, received signals, summaried signals, as well as the behavioural model, emission model, receiver model, and summarisation model are instances of the RDF/OWL classes defined in the signal ontology. As our ontology is defined using RDF/OWL standards, existing tools are available to support users to extend and edit the graph, such as Protégé (open source) and TopBraid Composer. A simplified view of the knowledge graph constructed is presented in \autoref{fig:signal-kg-instances}. The knowledge graph also includes probabilities (not shown), for example, a low prior probability that an employee will drop a tray.

\begin{figure}[tpb]
    \centering
    \includegraphics[width=0.9\linewidth]{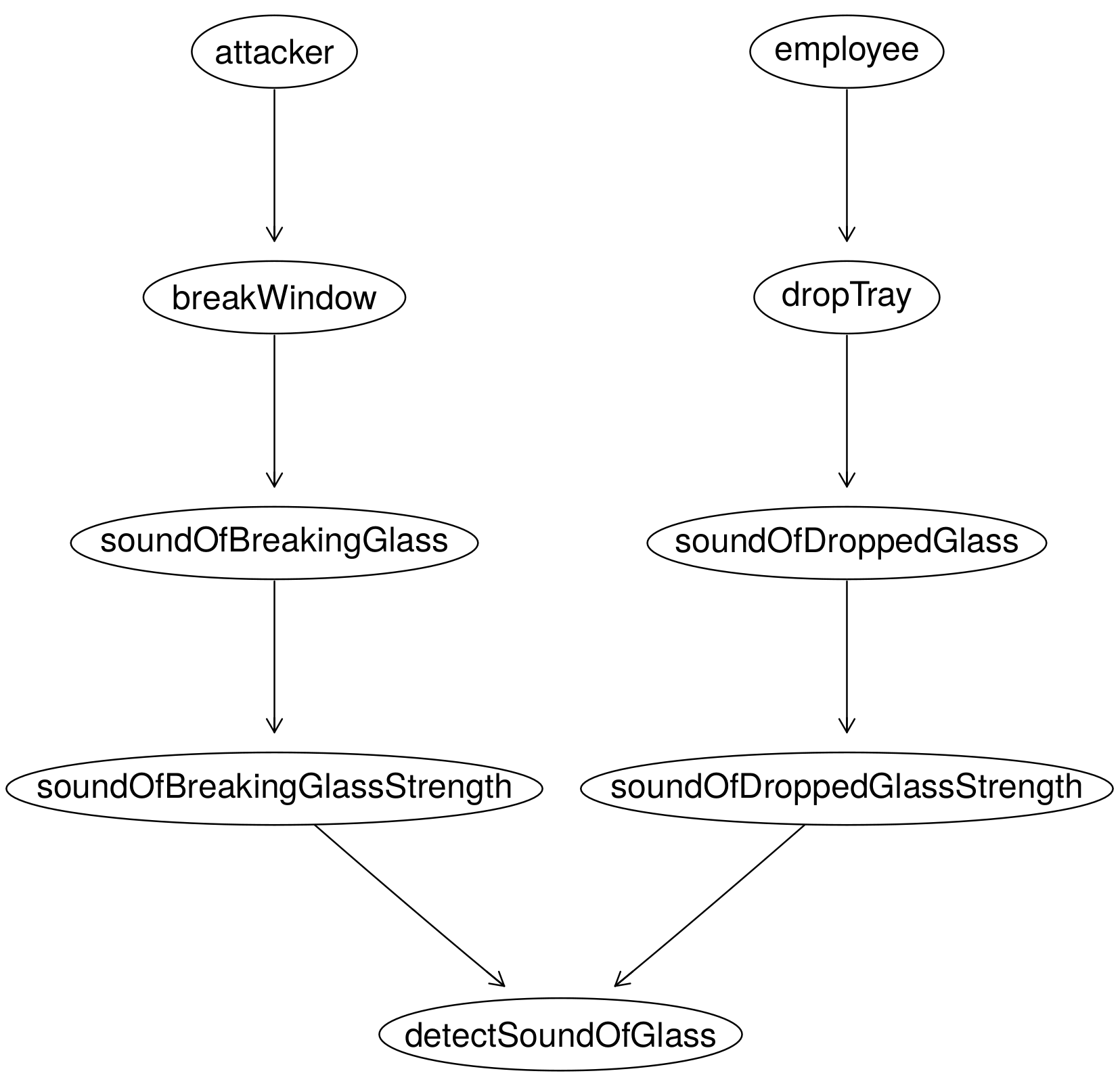}
    \caption{Generated Bayesian network for sound/audio}
    \label{fig:bayesian-network}
\end{figure}

The information in the signal knowledge graph along with the model of sensor placement within the building is then automatically converted to a causal graph by the signal reasoning engine. Our prototype implementation of the signal reasoning engine uses a Bayesian network to represent the causal graph, with nodes for each entity, action, and signal capable of being detected by a sensor. To ensure the signal reasoning engine can be reused for other signal types, the implementation of the signal reasoning engine only makes use of concepts in the signal ontology, not any particular signal defined in the signal knowledge graph. The generated Bayesian network for inferring the probability of an attacker given an observed sound is presented in \autoref{fig:bayesian-network} (conditional probability tables for each node are also generated, but are not shown for space reasons). While in this scenario the mapping between nodes in the Bayesian network and instances in our knowledge graph was 1:1, in the case of actions that can take place at multiple locations, or sensors within range of multiple signals, there can be a large number of nodes generated.

\subsection{Vision-Based Signals}

Vision based signals can be modelled in a similar manner to audio signals. However, they have different characteristics, for example, a sound-proof window blocks sound but not light, and while the intensity of a sound signal reduces with distance according to an inverse square law ($\propto1/d^2$) the size of objects in vision is proportional to the inverse of distance ($\propto1/d$). As an example of a vision-based signal, we consider detecting a knife in camera vision footage. Similar to the sound scenario in which the sound of glass is not necessarily suspicious, a knife is not suspicious if it is being used for cooking/eating in a dining area rather than as a weapon.

\subsection{Signals from Social Sensors}

As an example of signals from social sensors, we consider the case of detecting alarming tweets on Twitter at the location of interest. In contrast to sound/audio and vision, signals on social media arise as the result of a two stage process. Firstly, a bystander has to observe that something suspicious is occurring. This first stage of the process is similar to vision-based signals in which the bystander needs to be nearby to make the observation. The bystander themselves can be thought of as a sensor, albeit we cannot directly observe their thoughts. The bystander may then chose to tweet this information, which triggers the second stage of the process, in which the tweet is emitted as a digital signal, and may be observed by Twitter monitoring software which processes the stream of tweets for a particular location or set of criteria and classifies them as alarming or not. Based on the final classification we attempt to infer whether the underlying cause was an attacker. In addition to the possibility of a false alarm for non-malicious reasons, e.g., the Twitter monitoring software misinterprets a benign tweet as alarming, pranksters may generate prank tweets that look similar to good-faith alarming tweets.

\begin{figure}[tpb]
    \centering
    \includegraphics[width=\linewidth]{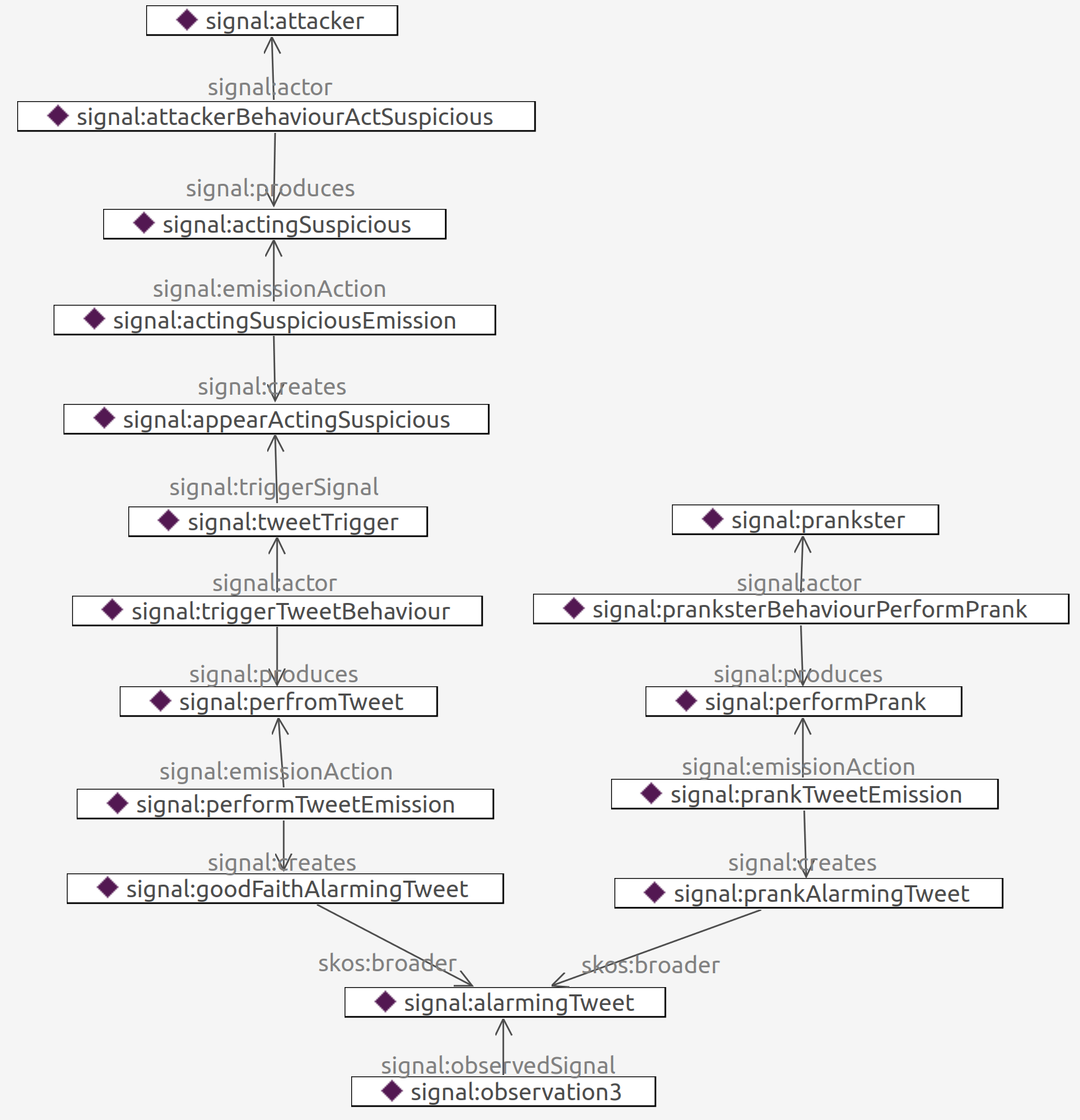}
    \caption{Partial view of knowledge graph created for social sensors}
    \label{fig:signal-kg-instances-tweets}
\end{figure}

\begin{figure}[tpb]
    \centering
    \includegraphics[width=\linewidth]{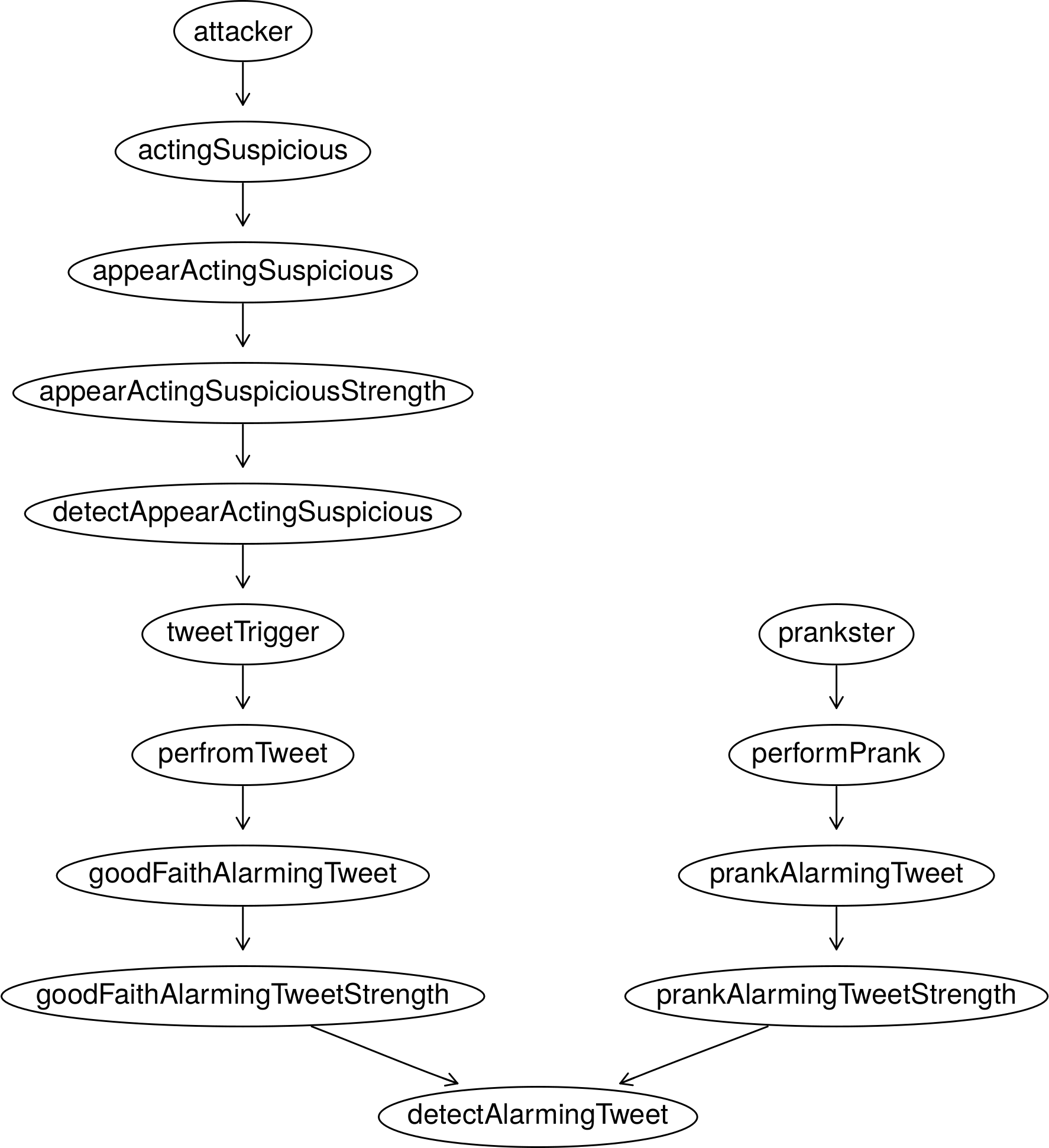}
    \caption{Generated Bayesian network for social sensors}
    \label{fig:bayesian-network-tweets}
\end{figure}

A simplified view of the knowledge graph constructed for signals from social sensors is presented in \autoref{fig:signal-kg-instances-tweets}. The corresponding Bayesian network generated by the signal reasoning engine for this scenario is presented in \autoref{fig:bayesian-network-tweets}.

\subsection{Integration}

\begin{figure*}[tpb]
    \centering
    \includegraphics[width=\linewidth]{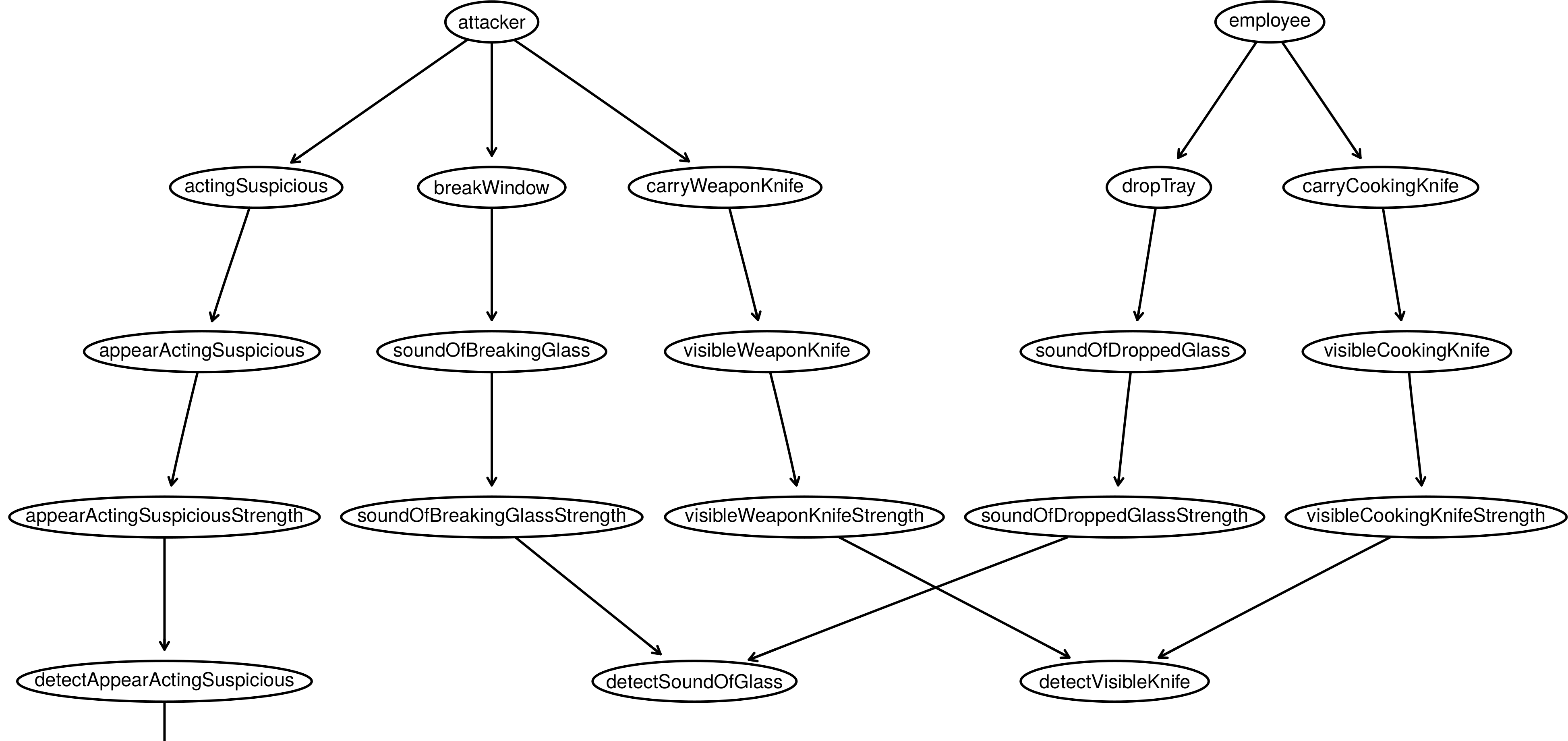}
    \caption{Part of generated Bayesian network integrating multiple types of signals}
    \label{fig:bayesian-network-full}
\end{figure*}

Sound/audio signals, vision-based signals, and signals from social sensors are all described using a common signal ontology, and form part of the same signal knowledge graph. This graph can be refined with additional causes of signals, or extended with new types of signals collected by different forms of sensors. Observations can be linked to the signal knowledge graph to support reasoning about the cause of the observations. A model of the building/space describing the placement of sensors and their surrounding environment is used to reason about the strength of signals at a particular location (e.g., if a sensor is far from a signal source, then it is unlikely to result in a detection) as well as actions that occur (e.g., a dropped tray of glasses is most likely to happen in the dining area). The signal reasoning engine combines this information to generate a single causal graph (specifically a Bayesian network, displayed in \autoref{fig:bayesian-network-full}), that can be used to infer the posterior probability of potential causes given all available sensor observations.









\section{Next Steps}
\label{sec:next-steps}

This section outlines research gaps that we plan to address in future research.

\subsection{Extension of Signal Knowledge Graph}

As outlined in \autoref{sec:existing-knowledge-graphs}, existing knowledge graphs contain limited information to reason about signals. Our approach in \autoref{sec:approach} presents the high level concepts that are needed, and \autoref{sec:demonstration} provides a sketch of what such a knowledge graph would look like. Future work is needed to create higher fidelity models and to grow the knowledge graph with additional actions that produce signals as well as new kinds of signals. This will likely require a crowd sourcing effort, or a way to automatically mine this information from unstructured sources. Furthermore, we intend to align our ontology with concepts in existing knowledge bases so that existing structured knowledge can be leveraged for reasoning.

\subsection{Generation of Probabilistic Programs for Reasoning over Sensor Data}

More accurately modelling signal propagation and sensor characteristics requires probabilistic functions over continuous space. To overcome the limitations of Bayesian networks, we intend to explore generation of probabilistic programs to achieve this task.

\subsection{Modelling the Characteristics of Machine Learning Classification Models}

Documentation for machine learning models usually includes overall performance metrics such as sensitivity and specificity. However, when machine learning models are used to classify a signal, it is important to capture how performance metrics change as a function of the signal strength. For example, to understand at what distance a sound/audio detection device making use of a classification model is able to detect different types of noises. Similarly, for vision-based signals, how large objects need to appear in the camera to be reliably recognised by an object recognition model. An improvement in classification performance metrics does not necessarily improve alarm generation, for example, if a sound/audio detection device can detect quiet sounds at long range, it could lead to false alarms due to detection of activities occurring outside the area of interest.

Previous research shows that the relationship between image size/resolution and classification performance of deep learning models is non-trivial and dependent on the neural network architecture and application domain. For example, \citet{Sabottke2020} examined the effect of image resolution on deep learning performance for the purpose of diagnosing radiology images and found that performance plateaus at resolutions of around 256x256 to 320x320 pixels, depending on the type of diagnosis. Interestingly, the performance classification for hernia appears to decrease at larger image resolutions. \citet{Richter2021} examined the effect of input size on CNN classifiers and found that the performance of the CNN is different when an input image is scaled to larger sizes despite not adding any new information.

\subsection{Modelling Adversarial Behaviour}

Our initial implementation infers the presence of an attacker given a set of signals at a particular time instant (or in the case of short pulse-like signals, over a time window). However, a better inference can be made by considering the full history/sequence of signals that arise from the attacker's behaviour and intent. For example, the opening of an external door, followed by footsteps through a hall toward a particular room. A challenge of modelling this is the adversarial nature of an attacker's movements---an attacker may deliberately take an indirect route in order to avoid detection or inference of their intent. One approach for this is to assume that the attacker picks the path that minimises their probability of detection \cite{simmons2022reliability}.

\subsection{Modelling Uncertainty}

Our approach allows for assigning probabilities, e.g., the probability that breaking a window will result in the sound of breaking glass. However, in some cases the probability will be unknown, in which case there needs to be a way to model the uncertainty of this parameter itself as a distribution. Furthermore, approaches are needed to learn (automatically calibrate) these parameters from data, suggest the optional sequence of actions/experiments to help the system calibrate itself, and to model probing behaviour of an attacker with partial knowledge of the environment attempting to learn more about it.


\subsection{Creation of Real World Signal Datasets}

Testing our approach requires real-world datasets of signal observations from multiple types of sensors recording simultaneously. While there exists large-scale image datasets of everyday scenes, e.g., Common Objects in Context (Microsoft COCO) \cite{Lin2014}, real-world datasets of other signals types such as sound/audio are limited. Existing datasets for sound detection include Audio Set \cite{Gemmeke2017}, a corpus of 527\footnote{The original Audio Set paper describes an ontology of 632 different classes of sounds, but some classes were excluded from labelling due to being obscure, abstract or difficult to find. The current release of Audio Set contains 527 classes of annotated sounds \url{https://research.google.com/audioset/}. However, some of these classes may lead to fairness/offence issues, thus Audio Set classification models such as YAMNet only use 521 classes \url{https://github.com/tensorflow/models/tree/master/research/audioset/yamnet\#class-vocabulary}.} different classes of annotated sounds in Youtube videos, Environmental Sound Classification (ESC) \cite{Piczak2015} consisting of 5-second snippets from Freesound and UrbanSound8K \cite{Salamon2014} which is also derived from Freesound. However, there are limited datasets of continuous sound/audio recordings; the CHiME-Home dataset \cite{Foster2015} records sounds in a domestic environment, but is limited in scope to a single sensor in one home.

To fill this gap we intend to capture our own dataset of observations from multiple types of sensors in a real-world environment, and represent the captured data in a form that can be linked with the signal knowledge graph to support reasoning. As it is unlikely that any threats will occur naturally in our dataset, we plan to also generate a synthetic dataset of signals that will occur under potential attack scenarios (e.g., the sound of a smashed glass, footsteps, then searching through drawers).

\section{Related Work}
\label{sec:related-work}

Existing literature covers systems and algorithms to fuse particular types of sensor data, use of knowledge graphs and deterministic rules to infer situational awareness, and use of probabilistic approaches to infer situational awareness. However, existing systems for inferring situational awareness do not provide a way to leverage knowledge graphs and probabilistic approaches together in an extensible way. Furthermore, they do not explicitly consider and model the nature of signals.

\subsection{Fusing Particular Types of Sensor Data}

\citet{Zieger2009a} present a system for detecting intrusions using a network of microphones (each in pairs to assist in determining the location of the sound source). To locate the source, they compute the Global Coherence Field (GCF) over all pairs of microphones to estimate the position of the sound source.
They tested the system through an intrusion scenario that involved asking people (acting as the intruder) to look through drawers in room for an object while trying to avoid detection. They also tested for false alarms through playing sounds such as train noises on a speaker outside the room. The alarm generation logic is based on the inferred sound location and a set of extracted sound features such as energy and spectral variation. However, their system used seven pairs of microphones to monitor a single room. They do not provide an approach to integrate other forms of sensor data or contextual information into the alarm generation logic.


\subsection{Knowledge Graphs and Deterministic Rules}

\citet{Roy2012} demonstrate the use of OWL DL (description logic) to reason about threats based on their attributes (intent, capability, opportunity), using illegal fishing as an example. However, they note that even for a simple example, they run into limitations of the reasoning system. Furthermore, they assume that threats can be classified into classes according to deterministic rules, whereas in reality threats may be probabilistic in nature.

\citet{Pai2017} demonstrate use of ontology based information fusion for situational awareness in military scenarios. Their approach allows for reasoning rules to infer which side observed units are on and what their status is, for example, based on what equipment they are observed to be using and whether they are advancing towards other units. However, these reasoning rules are deterministic rather than probabilistic.

\subsection{Probabilistic Rules}

\citet{Rogova2020} consider the scenario of a vehicle belonging to a known terrorist subject attacking a building. The paper presents a high level architecture for reasoning about such an attack, and suggests that argument mining or Concept Net \cite{Speer2017} could be used to automatically form hypotheses; however, it is unclear to what degree the hypothesis generation aspect was implemented. \citet{Ilin2021} realises the probabilistic part of the architecture.
A Bayesian network is used for detecting the threat, which considers opportunity, capability, intent, and matching of the suspect to the vehicle as factors. However, the Bayesian network and conditional probability tables need to be manually specified. As attacks are rare events, \citeauthor{Ilin2021} advocates for use of Choquet expected utility (which involves distorted probabilities similar to the way that humans perceive risk) instead of traditional expected utility to determine whether to flag potential attacks.

\citet{Yao2022} present an ontology for describing measurements, status, and tasks of an unmanned underwater vehicle (UUV). Their situational awareness approach makes use of both deterministic and probabilistic rules to reason about threats. Deterministic rules are used to identify potential threats in the environment, for example, when a target falls within a threshold distance it is flagged as a threat. In contrast, the impact of obstacles, energy margin, leakage points, abnormal task load, thruster, and steering gear on threat probability is computed probabilistically using a Bayesian network. The authors make use of BayesOWL \cite{Ding2004} to generate the Bayesian network from a probabilistic extended OWL ontology file. In theory, this approach provides the flexibility to specify probabilistic aspects in a similar form to deterministic rules; however, the probabilistic rules describing the Bayesian network still have to be explicitly specified (via OWL) rather than being automatically generated from other information in the knowledge graph.


\section{Conclusion}
\label{sec:conlustion}

Early identification of threats requires reasoning probabilistically over diverse sources of information. As attacks are rare events, reasoning systems need to leverage existing prior knowledge (in the form of a knowledge graph) to make an inference rather than relying solely on pattern matching and anomaly detection. In particular, there is a need to consider the potential causes, signal propagation and sensor characteristics to avoid false positives. However, existing knowledge graphs lack key information needed to support reasoning about signals. This paper lays the foundation for constructing a knowledge graph to fill this gap. Future work is also needed to collect larger real world datasets for tuning and testing signal reasoning systems.

\section*{Acknowledgment}

This paper was supported by research funding from the National Intelligence Postdoctoral Grant program (NIPG-2021-006).

Images of broken windows and glass used in this paper are licensed under CC BY 2.0 and available from \url{https://www.flickr.com/photos/frumbert/2835610296/} and \url{https://www.flickr.com/photos/machu/362660051/} respectively.

\bibliographystyle{IEEEtranN}
\bibliography{refs}

\end{document}